\documentclass[conference,a4paper]{IEEEtran}
\IEEEoverridecommandlockouts

\usepackage[hidelinks]{hyperref}
\usepackage[cmex10]{amsmath}%American Math Society(AMS) math formatting
\usepackage{amssymb,amsfonts}%AMS extra symbols and fonts
\usepackage{dblfloatfix}%fix double column figure ordering and placement

\usepackage[ruled,vlined]{algorithm2e}
\usepackage{multirow}
\usepackage{graphicx}
\usepackage[table]{xcolor}
\usepackage[font=footnotesize,labelfont=bf]{caption}

\usepackage{subcaption}
\usepackage{booktabs}
\usepackage[export]{adjustbox}
\usepackage{float}
\graphicspath{{Figures/PDF/}{Figures/PNG/}}

\usepackage{booktabs}
\usepackage{siunitx}
\usepackage[numbers,compress]{natbib}
\usepackage{texnames}
\usepackage{bm,bbm}
\usepackage{orcidlink}
\usepackage{xcolor}

\usepackage{booktabs}
\usepackage{geometry}
\usepackage{array}
\usepackage{graphicx}
\begin{document}

\title{Reconstruction Guided Few-shot Network For Remote Sensing Image Classification}

\author{
Mohit Jaiswal\textsuperscript{1}\orcidlink{0009-0002-1325-6892}, 
Naman Jain\textsuperscript{1}\orcidlink{0009-0007-1831-3426}, 
Shivani Pathak\textsuperscript{1}\orcidlink{0009-0006-6855-1134}, 
Mainak Singha\textsuperscript{2}\orcidlink{0000-0002-7615-2575}, \\
Nikunja Bihari Kar\textsuperscript{1}\orcidlink{}, 
Ankit Jha\textsuperscript{1}\orcidlink{0000-0002-1063-8978}, 
Biplab Banerjee\textsuperscript{2}\orcidlink{0000-0001-8371-8138} \\
\textsuperscript{1}\textit{Dept. of CSE, The LNMIIT Jaipur} \quad
\textsuperscript{2}\textit{CSRE, IIT Bombay} 
}

\maketitle
\begin{abstract}
Few-shot remote sensing image classification is challenging due to limited labeled samples and high variability in land-cover types. We propose a reconstruction-guided few-shot network \textit{(RGFS-Net)} that enhances generalization to unseen classes while preserving consistency for seen categories. Our method incorporates a masked image reconstruction task, where parts of the input are occluded and reconstructed to encourage semantically rich feature learning. This auxiliary task strengthens spatial understanding and improves class discrimination under low-data settings. We evaluated the efficacy of EuroSAT and PatternNet datasets under 1-shot and 5-shot protocols, our approach consistently outperforms existing baselines. The proposed method is simple, effective, and compatible with standard backbones, offering a robust solution for few-shot remote sensing classification. Codes are available at \url{https://github.com/stark0908/RGFS}.
\end{abstract}


\section{Introduction}

Remote sensing (RS) technology plays a vital role in acquiring detailed information about the Earth's surface and has witnessed significant advancements over recent decades. It is extensively applied across diverse domains including agricultural monitoring, environmental assessment, urban development, and disaster management \cite{yu2023deep,kaul2023literature}. Among the various RS data sources, aerial imagery offers high-resolution, localized observations that enable fine-grained analysis of geographic regions. Traditionally, classification of remote sensing data, particularly multispectral and hyperspectral images, has relied on classical machine learning algorithms such as Support Vector Machines (SVMs) \cite{melgani2004classification}, Random Forests (RF) \cite{pal2005random}, and k-Nearest Neighbors (k-NN) \cite{tarabalka2010spectral}. These conventional techniques depend heavily on large labeled datasets and handcrafted feature engineering to achieve satisfactory results. While effective for well-structured and annotated data, these methods often struggle to cope with complex spatial-spectral patterns, class imbalance, and limited supervision, thereby hindering their scalability and robustness in real-world applications \cite{yu2023deep,ghamisi2017advanced}.

Recent advances in deep learning have revolutionized remote sensing by enabling the automated extraction of discriminative features directly from raw image data using convolutional neural networks (CNNs) and transformer-based models \cite{ball2017comprehensive,aleissaee2023transformers}. These models demonstrate superior performance, robustness, and scalability compared to traditional approaches. However, their effectiveness is frequently constrained by the need for large-scale annotated datasets, which are often costly or infeasible to obtain in practical settings \cite{ball2017comprehensive,li2020rs,yu2023deep}. Few-shot learning (FSL) addresses this limitation by enabling models to generalize from only a limited number of labeled samples per class \cite{li2020rs,sung2018learning,koch2015siamese,finn2017model,10286511}. FSL methods including Relation Networks \cite{sung2018learning}, Siamese Networks \cite{koch2015siamese}, and Model-Agnostic Meta-Learning (MAML) \cite{finn2017model} have been adapted successfully for remote sensing, facilitating rapid adaptation to novel classes with minimal supervision \cite{li2020rs}. More recently, Stable Prototypical Networks (SPN) \cite{pal2021spn} have enhanced prototypical networks by introducing variance reduction techniques that improve prototype stability and generalization across diverse tasks \cite{li2020rs}. These advancements underscore the importance of learning stable and discriminative feature representations to achieve reliable few-shot classification on complex remote sensing datasets.

Motivated by the limitations observed in SPN, we propose a reconstruction-guided regularization framework integrated into our model architecture. This approach simultaneously optimizes the classification objective and an image reconstruction loss by decoding latent embeddings. The use of masked reconstruction compels the model to infer missing image patches, effectively encouraging the capture of contextual dependencies. Our proposed RGFS-Net incorporates masked reconstruction loss within a meta-learning framework to facilitate robust adaptation in few-shot classification scenarios. This encourages the encoder to preserve meaningful spatial-spectral structures, enhancing feature consistency and discriminability across tasks. Additionally, we explore architectural enhancements to further refine the latent feature space. We summarize our contributions as follows:
\begin{itemize}
    \item We introduce RGFS-Net, a reconstruction-guided few-shot learning framework that incorporates masked image reconstruction loss to enhance feature robustness and stability in remote sensing classification.
    \item We design a learnable bottleneck layer between the encoder and decoder, enforcing compact and discriminative latent embeddings, which improves classification performance on both seen and unseen classes.
    \item We conduct extensive experiments on benchmark datasets, EuroSAT and PatternNet, demonstrating that RGFS-Net outperforms existing few-shot learning baselines in terms of classification accuracy and consistency.
\end{itemize}
\begin{figure*}[ht!]
    \centering
    \includegraphics[width =\textwidth]{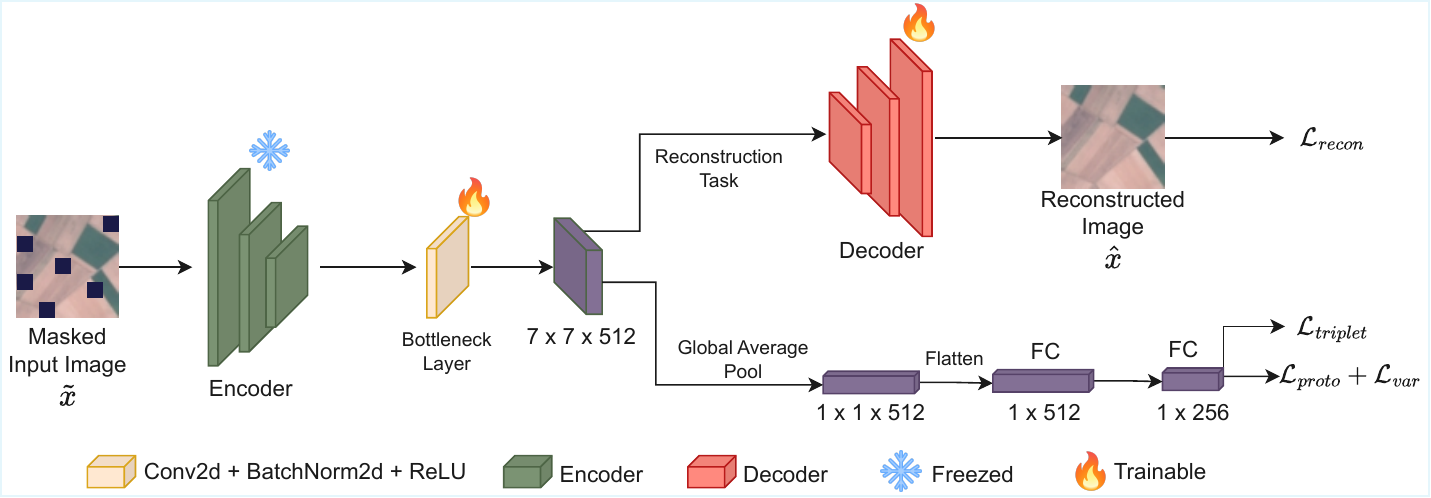}
    \caption{The model architecture of our proposed RGFS-Net. Here, we extract the image features from the pre-trained image encoder like (VGG-Net) which we have kept frozen during the training. We perform the reconstruction of masked patched image using learnable decoder under the supervision of reconstruction loss and mask loss. In order to better segregate the representaion space, we use the triplet loss along with the cross entropy loss.}
    \label{figure}
\end{figure*}

\section{Methodology and Proposed Model}
\label{sec:method_proposed_model}

\subsection{Problem formulation}
Let $\mathcal{D} = \{\mathcal{D}_\text{base}, \mathcal{D}_\text{novel}\}$ represent the remote sensing dataset, where $\mathcal{D}_\text{base} = {(x_i, y_i)}_{i=1}^{N_\text{base}}$ contains labeled samples from base classes $\mathcal{C}_\text{base}$ used for training, $\mathcal{D}_\text{novel} = {(x_j, y_j)}_{j=1}^{N_\text{novel}}$ contains limited labeled samples from novel classes $\mathcal{C}_\text{novel}$ for few-shot evaluation, with $\mathcal{C}_\text{base} \cap \mathcal{C}_\text{novel} = \emptyset$.
Each input $x \in \mathbb{R}^{H \times W \times C}$ is a remote sensing image with height $H$, width $W$, and $C$ spectral or RGB channels, and $y \in \mathcal{C}$ is the corresponding class label.

\subsection{The Proposed RGFS-Net}
In Figure \ref{figure}, we present the architecture overview of RGFS-Net, which integrates a self-supervised reconstruction task with an embedding generation mechanism, both stemming from a shared feature encoder. 
Let $\mathbf{x} \in \mathbb{R}^{H \times W \times C}$ be an input image, where $H$, $W$, and $C$ represent the height, width, and number of channels, respectively. A structured block-wise mask $\mathbf{M} \in \{0, 1\}^{H \times W}$ is first applied, generating a masked image $\tilde{\mathbf{x}} = \mathbf{x} \odot (\mathbf{1} - \mathbf{M})$, where $\odot$ denotes the Hadamard product. The encoder, $f_{\text{enc}}$, which is built upon a pre-trained CNN backbone, processes this masked input to produce a single, rich latent representation $\mathbf{z} \in \mathbb{R}^{h \times w \times d}$ \cite{He_2016_CVPR}. This latent space $\mathbf{z}$ serves as a crucial junction for two simultaneous pathways. The first is the reconstruction pathway, where a decoder, $g_{\text{dec}}$, composed of successive transposed convolutional layers, upsamples $\mathbf{z}$ to generate a reconstructed image $\hat{\mathbf{x}}$. The network is trained by minimizing the reconstruction loss, $\mathcal{L}_{\text{recon}} = \|\mathbf{x} - \hat{\mathbf{x}}\|_2^2$, which forces the encoder to capture high-level semantic information necessary to infer the missing content[7]. The second pathway is the embedding pathway. An embedding head, $h_{\text{emb}}$, which consists of an adaptive average pooling layer followed by a fully connected (FC) layer, also takes $\mathbf{z}$ as input and projects it into a final, compact feature vector $\mathbf{e} \in \mathbb{R}^{D}$ suitable for downstream tasks[8].

This unified design ensures that the embedding vector $\mathbf{e}$ is derived from features that are proven to be effective for holistic image reconstruction. The entire model, comprising the encoder, decoder, and embedding head, is trained end-to-end during the self-supervised pre-training phase. For the final few-shot classification task, the decoder is discarded, and the pre-trained encoder and embedding head are used as a powerful, fixed feature extractor. For spatial regularization and robust the latent space, we integrate DropBlock \cite{NEURIPS2018_7edcfb2d} layers into the encoder's backbone. By nullifying contiguous regions of feature maps, this technique discourages the network from relying on specific local activations and promotes the learning of more generalized spatial representations.

\subsection{Training methodology}
Our training procedure follows the episodic learning paradigm inspired by Prototypical Networks \cite{NIPS2017_cb8da676}, but extends it with a multi-task objective tailored to encourage robust, stable, and transferable representations. During each training iteration, an episode is constructed by sampling a support set $\mathcal{S}$ and a query set $\mathcal{Q}$. Simultaneously, a separate set of images is sampled for the self-supervised reconstruction task. A key component of our approach is the integration of stochastic forward passes, drawing inspiration from Monte Carlo sampling techniques \cite{pmlr-v48-gal16}. Specifically, for each episode, we perform $n$ stochastic passes through the model, enabled by regularization mechanisms such as DropBlock, which introduce variability in the activations. This results in $n$ distinct sets of embeddings for the support and query samples, enabling the approximation of a posterior distribution over the model outputs.

During the forward pass, class prototypes are computed by averaging the support embeddings of each class. Simultaneously, the masked versions of the images are passed through the decoder to reconstruct the original inputs, supervised via a self-supervised reconstruction objective. This dual-task setup ensures that the encoder learns features that are not only discriminative for classification but also rich enough to support spatial-spectral reconstruction. The overall training objective is a composite loss function comprising four components: classification loss, reconstruction loss, structure-preserving regularization and a consistency loss that penalizes variance across the $n$ stochastic predictions for each query sample. By explicitly minimizing this variance for the true class, the model is encouraged to make confident and stable predictions, reducing epistemic uncertainty and enhancing generalization in FSL settings.

\subsection{Objective function}
The RGFS-Net is trained end-to-end using a multi-task objective that integrates four complementary loss components: a prototypical classification loss, a metric learning-based triplet loss, a masked reconstruction loss, and a variance regularization loss. Together, these terms guide the model to learn discriminative, robust, and generalizable representations for few-shot classification.

\textbf{1) Prototypical classification loss:}  
For each of the $n$ stochastic forward passes, we compute the prototypical loss following \cite{NIPS2017_cb8da676}. Given a feature encoder $f_{\phi}$ with parameters $\phi$, the prototype $\mathbf{c}_k$ for class $k$ is computed as the mean of support embeddings. The probability that a query sample $\mathbf{x}_i$ belongs to class $k$ is obtained via a softmax over the negative squared Euclidean distances to the prototypes. The classification loss is then averaged over $n$ passes:
\begin{equation}
\small
\begin{aligned}
\mathcal{L}_{\text{proto}} = \frac{1}{n} \sum_{j=1}^{n} 
\mathbb{E}_{(\mathbf{x}_i, y_i=k) \in \mathcal{Q}} \bigg[ 
& -\log \bigg( \frac{ e \left( -d(f_{\phi}^{(j)}(\mathbf{x}_i), \mathbf{c}_k^{(j)}) \right) }
{\sum\limits_{k'} e \left( -d(f_{\phi}^{(j)}(\mathbf{x}_i), \mathbf{c}_{k'}^{(j)}) \right) } \bigg)
\bigg]
\end{aligned}
\label{eq:proto_loss}
\end{equation}

where $f_{\phi}^{(j)}$ and $\mathbf{c}_k^{(j)}$ denote the encoder output and class prototype in the $j$-th pass, and $d(\cdot, \cdot)$ is the squared Euclidean distance.

\textbf{2) Triplet margin loss:}  
To further refine the embedding space, we use a triplet margin loss that pulls query embeddings closer to their correct class prototype (positive) and pushes them away from the nearest incorrect prototype (hard negative). The loss is averaged across all passes:
\begin{equation}
\small
\begin{aligned}
\mathcal{L}_{\text{triplet}} = \frac{1}{n} \sum_{j=1}^{n} 
\mathbb{E}_{(\mathbf{x}_i, y_i=k) \in \mathcal{Q}} \Big[
& \max \big( 0, m + d(f_{\phi}^{(j)}(\mathbf{x}_i), \mathbf{c}_{k}^{(j)}) \\
& \quad - d(f_{\phi}^{(j)}(\mathbf{x}_i), \mathbf{c}_{k'}^{-(j)}) \big) 
\Big]
\end{aligned}
\label{eq:triplet_loss}
\end{equation}

where $m$ is the margin, and $\mathbf{c}_{k'}^{-(j)}$ is the closest incorrect prototype in the $j$-th pass.

\textbf{3) Reconstruction loss:}  
To promote spatial consistency and contextual reasoning, we employ a self-supervised reconstruction loss that combines masked and global L1 losses:
\begin{equation}
\begin{split}
    \mathcal{L}_{\text{recon}} = \frac{1}{n} \sum_{j=1}^{n} \left( 
    \|\mathbf{M} \odot (\mathbf{x} - \hat{\mathbf{x}}^{(j)})\|_1 + \|\mathbf{x} - \hat{\mathbf{x}}^{(j)}\|_1 \right)
\end{split}
\label{eq:recon_loss}
\end{equation}
where $\mathbf{M}$ is a binary mask, $\hat{\mathbf{x}}^{(j)}$ is the reconstructed image in the $j$-th pass, and $\odot$ is the Hadamard product. This loss enforces faithful recovery of masked regions while preserving the overall image structure.

\textbf{4) Variance regularization loss:}  
To improve prediction stability under stochastic regularization, we penalize the variance in predicted class probabilities for each query image across multiple passes, similar to SPN \cite{pal2021spn}.
\begin{equation}
\begin{split}
    \mathcal{L}_{\text{var}} = \sum_{(\mathbf{x}_i, y_i=k) \in \mathcal{Q}} \sqrt{\frac{1}{n} \sum_{j=1}^{n} (p_{i,k}^{(j)} - \bar{p}_{i,k})^2}
\end{split}
\label{eq:var_loss}
\end{equation}
where $p_{i,k}^{(j)}$ is the predicted probability of the true class $k$ for query $\mathbf{x}_i$ in the $j$-th pass, and $\bar{p}_{i,k}$ is the corresponding mean over all $n$ passes.

\textbf{Total Loss:}  
The final training objective is a weighted sum of all the components, with   $\alpha$ , $\beta$ and $\lambda$ are hyperparameters that balance the contributions of reconstruction and variance regularization, respectively.
\begin{equation}
    \mathcal{L}_{\text{total}} = \mathcal{L}_{\text{proto}} + \alpha \cdot \mathcal{L}_{\text{var}} + \beta \cdot \mathcal{L}_{\text{triplet}} + \lambda \cdot \mathcal{L}_{\text{recon}}
\label{eq:total_loss}
\end{equation}

\section{Experimental Evaluation}
\label{sec:exp}
\subsection{Dataset description}
We conduct experiments on two benchmark remote sensing datasets: i) \textbf{EuroSat:} This dataset consists of 27,000 RGB satellite images from Sentinel-2, spanning 10 land-use classes such as Forest, Industrial, River, and Residential. Each image has a spatial resolution of 10 meters and a size of 64×64 pixels. Its rich seasonal and geographical diversity makes it suitable for few-shot and domain generalization tasks. ii) \textbf{PatternNet:} Comprising 24,000 high-resolution RGB images across 30 scene categories including Bridge, Baseball Field, Parking Lot, and Windmill, each class has 800 images of size 256×256, with spatial resolutions ranging from 0.062 to 4.693 meters. The dataset presents strong inter-class similarity, offering a challenging few-shot classification benchmark. We adopt 1-shot and 5-shot evaluation settings with base/novel class splits of 19/19 for PatternNet and 5/5 for EuroSat. Experimentally we found $\alpha$= 0.01, $\beta$=1 and $\lambda$=5 to give the best performance. The margin for the triplet loss is 1.5. To assess generalization, we report results separately for both seen and unseen class subsets.

\subsection{Comparison to the literature}

We compare RGFS-Net with state-of-the-art (SOTA) few-shot learning (FSL) methods including Siamese Networks \cite{koch2015siamese}, Relation Networks \cite{sung2018learning}, Prototypical Networks \cite{NIPS2017_cb8da676}, and Stable Prototypical Networks (SPN) \cite{pal2021spn}, across multiple way-shot configurations on the EuroSAT and PatternNet datasets. The comparative results are summarized in Tables~\ref{tab:eurosat} and~\ref{tab:patternet}. On the EuroSAT dataset, RGFS-Net demonstrates consistent improvements over existing methods, particularly in more challenging 5-way settings. For instance, in the 5-way 5-shot configuration, our model achieves an unseen accuracy of 85.64\%, outperforming SPN (70.85\%) by 14.79 percentage points. Similarly, in the 5-way 1-shot case, RGFS-Net achieves 58.48\% compared to SPN's 58.34\%. In the 3-way 5-shot setting, RGFS-Net slightly surpasses SPN by 1.8\% on unseen classes. Although SPN marginally outperforms our model by 4.71\% in the 3-way 1-shot scenario, RGFS-Net maintains better stability across broader configurations. On average, RGFS-Net improves upon SPN by approximately 5.4\% across all unseen settings on EuroSAT. We observe the similar trends on the PatternNet dataset. In the 5-way 5-shot configuration, RGFS-Net achieves an unseen accuracy of 95.37\%, improving upon SPN (86.87\%) by 8.5\%. Additionally, our model achieves 97.62\% and 88.83\% in the 3-way 5-shot and 3-way 1-shot settings, respectively, showing consistent gains. Averaged across all settings, RGFS-Net surpasses SPN by around 6.7\% on unseen class performance. These empirical results demonstrate that the reconstruction-guided regularization employed in RGFS-Net effectively enhances feature generalization in few-shot remote sensing classification. By leveraging masked reconstruction and a bottleneck-regulated latent space, the model better captures spatial-spectral dependencies and yields more stable representations under low-data regimes.

\begin{table*}[htbp]
\centering
\caption{Few-shot classification accuracy (\%) on EuroSAT dataset with training and testing splits of 5-5. Here, `All` indicates all classes.}
\footnotesize
\begin{tabular}{lcccccccc}
\toprule
\multirow{2}{*}{\textbf{Model}} 
& \multicolumn{2}{c}{3-way 1-shot} & \multicolumn{2}{c}{3-way 5-shot} 
& \multicolumn{2}{c}{5-way 1-shot} & \multicolumn{2}{c}{5-way 5-shot} \\
\cmidrule(lr){2-3} \cmidrule(lr){4-5} \cmidrule(lr){6-7} \cmidrule(lr){8-9}
& All & Unseen & All & Unseen & All & Unseen & All & Unseen \\
\midrule
Siamese \cite{koch2015siamese}   & 51.89 & 64.17 & --    & --    & 39.17 & 52.29 & --    & -- \\
RelationNet \cite{sung2018learning}& 80.64 & 68.44 & 86.47 & 75.98 & 69.57 & 50.46 & 74.34 & 60.31 \\
% MAML         & --    & --    & --    & --    & --    & --    & --    & -- \\
Prototypical \cite{NIPS2017_cb8da676}   & 83.68 & 70.08 & 88.33 & 73.43 & 72.06 & 47.13 & 81.29 & 61.71 \\
SPN \cite{pal2021spn}         & 84.71 & \textbf{72.11 }& 92.13 & 83.16 & 75.20 & 58.34 & 85.37 & 70.85 \\
\rowcolor{gray!15} 
RGFS-Net (Ours)         & \textbf{84.83} & 67.40 & \textbf{92.83} & \textbf{84.96} & \textbf{80.36} & \textbf{58.48} & \textbf{90.90} & \textbf{85.64} \\
\bottomrule
\end{tabular}
\label{tab:eurosat}
\end{table*}

\begin{table*}[htbp]
\centering
\caption{Few-shot classification accuracy (\%) on PatternNet dataset with training and testing splits of 19-19. Here, `All` indicates all classes.}
\footnotesize
\begin{tabular}{lcccccccc}
\toprule
\multirow{2}{*}{\textbf{Model}} 
& \multicolumn{2}{c}{3-way 1-shot} & \multicolumn{2}{c}{3-way 5-shot} 
& \multicolumn{2}{c}{5-way 1-shot} & \multicolumn{2}{c}{5-way 5-shot} \\
\cmidrule(lr){2-3} \cmidrule(lr){4-5} \cmidrule(lr){6-7} \cmidrule(lr){8-9}
& All & Unseen & All & Unseen & All & Unseen & All & Unseen \\
\midrule
Siamese \cite{koch2015siamese}   & 53.61 & 57.72 & --    & --    & 41.10 & 46.97 & --    & -- \\
RelationNet \cite{sung2018learning} & 88.62 & 76.72 & 92.53 & 83.99 & 87.13 & 73.95 & 91.87 & 83.65 \\
% MAML         & --    & --    & --    & --    & --    & --    & --    & -- \\
Prototypical \cite{NIPS2017_cb8da676}   & 90.98 & 76.34 & 94.24 & 85.98 & 87.86 & 66.48 & 91.32 & 81.11 \\
SPN \cite{pal2021spn}     & 91.07 & 76.34 & 97.12 & 92.10 & 87.27 & 68.73 & 95.13 & 86.87 \\
\rowcolor{gray!15}
RGFS-Net (Ours)         & \textbf{92.67}    & \textbf{88.83}    & \textbf{98.06 }   & \textbf{97.62}   &\textbf{88.32}    & \textbf{85.34 }  & \textbf{98.36} & \textbf{95.37} \\
\bottomrule
\end{tabular}
\label{tab:patternet}
\vspace{-0.3cm}
\end{table*}

\subsection{Ablation studies}
\noindent\textbf{Ablation to the architecture:}  
We ablate RGFS-Net by incrementally adding core components to isolate their contributions to overall and unseen class performance (Table~\ref{tab:ablation_components}). Starting with a baseline model trained using only cross-entropy loss, we observe 82.65\% accuracy on all classes and 65.30\% on unseen classes. Adding the reconstruction loss significantly improves generalization, increasing unseen accuracy by +9.53\% (to 74.83\%), with a minor gain in overall accuracy. Incorporating masked loss alongside reconstruction further enhances both metrics, unseen accuracy reaches 75.76\% and overall accuracy rises to 85.00\%. We ablate with triplet loss, which helps in refining the embedding space. Here, the setup yields 85.30\% overall and 76.38\% unseen accuracy, suggesting better class separability. Adding a bottleneck layer improves abstraction, boosting performance to 87.30\% and 77.68\% on all and unseen classes, respectively. Finally, our complete i.e., RGFS-Net integrated with all components achieves 90.90\% accuracy on all classes and 85.64\% on unseen classes, outperforming the prototypical classification $\mathcal{L}_{proto}$ only baseline by +8.25\% (all) and +20.34\% (unseen). These gains demonstrate the effectiveness of our design choices for robust few-shot classification.

\begin{table}[htbp]
\centering
\caption{Ablation study for loss components the on EuroSAT dataset for 5-way 5-shot setting. Here, w/ denotes with.}
\footnotesize
\begin{tabular}{lcc}
\toprule
\textbf{Component} & All & Unseen \\
\midrule
With $\mathcal{L}_{proto}+\mathcal{L}_{var}$(Baseline)                          & 82.65 & 65.30 \\
+ $\mathcal{L}_{recon}$                          & 82.86 & 74.83 \\
+ Mask Loss (w/ $\mathcal{L}_{recon}$) & 85.00 & 75.76 \\
+ $\mathcal{L}_{triplet}$                                  & 85.30 & 76.38 \\
+ Bottleneck Layer                               & 87.30 & 77.68 \\
\rowcolor{gray!15}
RGFS-Net                                & \textbf{90.90} & \textbf{85.64} \\
\bottomrule
\end{tabular}
\label{tab:ablation_components}
\end{table}

\begin{figure}[ht!]
    \centering
    \includegraphics[width=\columnwidth]{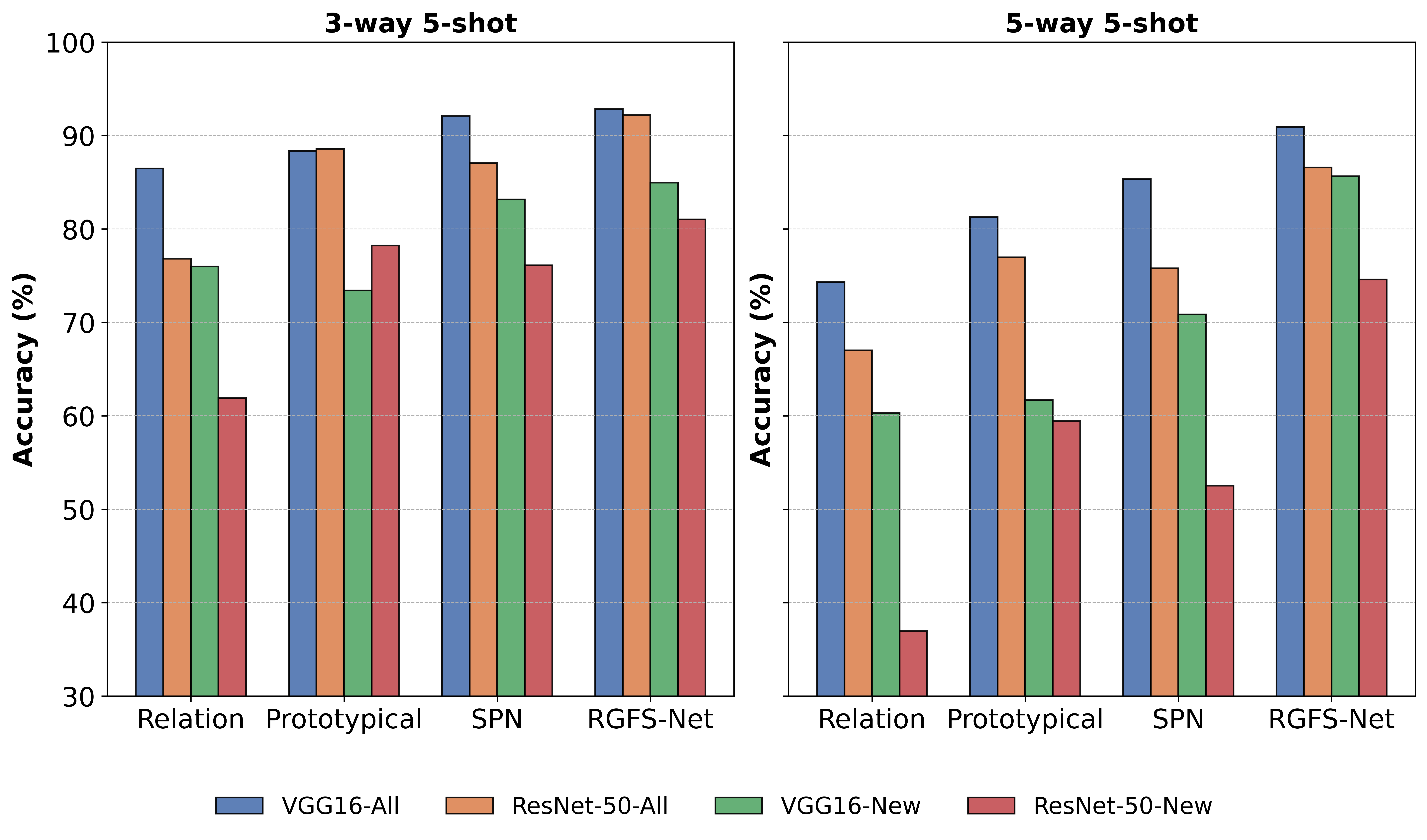}
    \caption{Comparison of model performance by varying the feature extractor backbone on the EuroSat dataset.}
    \label{fig:backbone}
\end{figure}

\noindent\textbf{Ablation with feature extractor backbone:} To evaluate the robustness and generalizability of RGFS-Net across different backbone architectures, we conduct experiments using two widely adopted pre-trained feature extractors: VGG16 and ResNet-50. As shown in Figure~\ref{fig:backbone}, RGFS-Net consistently outperforms existing few-shot learning baselines across both 3-way and 5-way settings, irrespective of the backbone used. These results not only demonstrate the effectiveness of our approach but also indicate that RGFS-Net is agnostic to the choice of feature extractor, maintaining superior performance across architectures.

\noindent\textbf{Ablation by varying training and testing splits:} In Table \ref{tab:split_ablation}, we experiment with various FSL methods on different training and testing splits with 3-way 1-shot setting on EuroSat dataset. We observe that the proposed RGFS-Net outperforms all SOTA methods by at least 4.25\% on the unseen classes. To establish the benchmark, we designated 5 classes as seen (training) and the remaining 5 classes as unseen (testing) for the experiments across all datasets.

\begin{table}[htbp]
\centering
\caption{Comparison of few-shot learning methods under varying train-test splits in the 3-way 1-shot setting on EuroSAT dataset. Accuracy (\%) is reported for all classes and Unseen classes.}
\footnotesize
\begin{tabular}{lcccccc}
\toprule
\multirow{2}{*}{\textbf{Method}} 
& \multicolumn{2}{c}{3 Train - 7 Test} 
& \multicolumn{2}{c}{5 Train - 5 Test} 
& \multicolumn{2}{c}{7 Train - 3 Test} \\
\cmidrule(lr){2-3} \cmidrule(lr){4-5} \cmidrule(lr){6-7}
& All & Unseen & All & Unseen& All & Unseen\\
\midrule
Relation      & 66.97 & 55.79 & 80.64 & 68.44 & 86.27 & 77.23 \\
Prototypical  & 73.62 & 62.80 & 83.68 & 70.08 & 91.28 & 76.24 \\
SPN           & 73.79 & 62.17 & 84.71 & \textbf{72.11} & 91.13 & 78.23 \\
\rowcolor{gray!15} RGFS-Net      & \textbf{80.30} & \textbf{70.76} & \textbf{84.83} & 67.40 & \textbf{95.23} & \textbf{79.23} \\
\bottomrule
\end{tabular}
\label{tab:split_ablation}
\end{table}

\section{Conclusion}
In this work, we introduced RGFS-Net, a reconstruction-guided few-shot learning setup designed for remote sensing image classification. We have integrated the masked reconstruction loss and a learnable bottleneck within a meta-learning setup, our model effectively captures rich spatial-spectral features that improve generalization to unseen classes. We perform extensive experiments on EuroSAT and PatternNet datasets demonstrate that RGFS-Net consistently outperforms the state-of-the-art few-shot methods. Our ablation studies further highlight the importance of our architectural choices and auxiliary losses in enhancing feature robustness and representation quality. Finally, we plan to extent our contribution to other remote sensing modalities and more complex real-world tasks, paving the way for more reliable and scalable few-shot classification in practical applications.

\small
\bibliographystyle{IEEEtranN}
\bibliography{references}

\end{document}